\pdfoutput=1
\documentclass[11pt,a4paper]{article}
\usepackage[hyperref]{emnlp2022}
\usepackage{times}
\usepackage{latexsym}

\usepackage{microtype}

\usepackage[utf8]{inputenc}
\usepackage[T1]{fontenc}
\usepackage{booktabs}
\usepackage{footnote}

\usepackage{amsmath}
\usepackage{amssymb}
\usepackage{comment}
\usepackage{graphicx}
\usepackage{caption, subcaption}
\usepackage{xcolor}

\usepackage{amsmath}
\usepackage{algpseudocode}

\usepackage{algorithm}

\usepackage{tikz}
\usepackage{xspace}
\usetikzlibrary{positioning}

\title{Boosting Natural Language Generation from Instructions\\ with Meta-Learning}

\author{
Budhaditya Deb, Guoqing Zheng, Ahmed Hassan Awadallah\\
Microsoft Research\\
\texttt{\{budeb, zheng, hassanam\}@microsoft.com} \\}
\date{}

\begin{document}
    \maketitle

\begin{abstract}
    Recent work has shown that language models (LMs) trained with multi-task \textit{instructional learning} (MTIL) can solve diverse NLP tasks in zero- and few-shot settings with improved performance compared to prompt tuning. MTIL illustrates that LMs can extract and use information about the task from instructions beyond the surface patterns of the inputs and outputs. This suggests that meta-learning may further enhance the utilization of instructions for effective task transfer. In this paper we investigate whether  meta-learning applied to MTIL can further improve  generalization  to unseen tasks in a zero-shot setting. Specifically, we propose to adapt meta-learning to MTIL in three directions: 1) Model Agnostic Meta Learning (MAML), 2) Hyper-Network (HNet) based adaptation to generate task specific parameters conditioned on instructions, and 3) an approach combining HNet and MAML. Through extensive experiments on the large scale Natural Instructions V2 dataset, we show that our proposed approaches significantly improve over strong baselines in zero-shot settings. In particular, meta-learning improves the effectiveness of instructions and is most impactful when the test tasks are strictly zero-shot (i.e. no similar tasks in the training set) and are "hard" for LMs, illustrating the potential of meta-learning for MTIL for out-of-distribution tasks.
    

\end{abstract}

\section{Introduction}

    Given some basic instructions and a few demonstrations, humans are capable of conducting diverse tasks without any supervision. Can language models perform similarly on unseen tasks when trained with instructions? Specifically can such an approach work on complex generation tasks with relatively smaller language models (LMs)?
    
    The recent advances in large LMs have shown tremendous potential in diverse AI applications and have the capability to change the way model developers and users interact with intelligent systems. The inherent representative power of such models has shown that diverse NLP tasks can be solved purely by appending prompts or demonstrations in context before a test input~\cite{Radford2019LanguageMA, GPT3}. This has led to the rise of prompt-based training~\cite{Liu2021PretrainPA} where even much smaller models trained on a large set of tasks in a multi-task setting with prompts, can behave similarly~\cite{DBLP:conf/naacl/SchickS21}.
    
    
    A natural extension of the prompt tuning concept involves adding instructions about the task along with the demonstrations. Instructions are more informative than prompts and aid the language models to solve unseen tasks better. Instructions can have different forms, for example to convey a short task specific statement (e.g. "Provide a short summary for the following input")~\cite{schick-schutze-2021-shot}, or a natural language question ("How would you rephrase that in a few words?")~\cite{sanh2022multitask, wei2022finetuned, Bach2022PromptSourceAI}. However, for complex generation tasks, short instructions can be ambiguous and non-informative and thus need large LMs which can encode a much richer prior knowledge. 
    
    In contrast,~\cite{WangMishra2022NaturalIB} define instructions in the Natural Instructions V2 (NIV2) dataset comprising of detailed task descriptions, positive and negative examples, and explanations. Instructions in NIV2 are similar to annotation guidelines, and thus potentially more beneficial\footnote{The instructions in NIV2 are in-fact taken from annotation guidelines for each of the tasks}. Using multi-task \textit{instructional learning} (MTIL) on diverse tasks,~\cite{WangMishra2022NaturalIB} showed that even smaller models can be competitive with larger models on zero-shot generalization to unseen tasks.

    Results in ~\cite{WangMishra2022NaturalIB} illustrated that LMs can extract useful information from instructions beyond the surface patterns available in the prompts for solving a task. This suggests that learning-to-learn or meta-learning paradigm can further enhance the utilization of instructions by learning about task at deeper levels. In this paper, we investigate  how smaller LMs could best benefit from the natural instructions and whether meta-learning paradigms can further improve the zero-shot generalization ability of LMs in MTIL. Meta-learning has been shown to be effective in adapting knowledge with little supervision but to the best of our knowledge has not been adapted to MTIL in zero-shot settings.
    
    
    
    
    
    Specifically, we explore two different meta-learning approaches. First we propose to adapt Model Agnostic Meta Learning (MAML)~\cite{Finn2017ModelAgnosticMF} for MTIL, an optimization based approach. Second, we explore hyper-network (HNet)~\cite{Ha2017HyperNetworks} based MTIL, a black-box approach. HNet introduces an auxiliary LM which encodes instructions to produce task specific parameters which are added to the main LM parameters to generate a task specific LM at prediction time. In addition, we evaluate a third approach which combines the two into a HNet-MAML by training the HNet model using MAML. 
    
    We conduct extensive experiments
    specifically designed to test the generalization ability of LMs trained with instructions under different zero shot conditions. We use two sets of training tasks from the NIV2 dataset: 1) all natural language tasks and 2) natural language generation tasks. We evaluate the models for two sets of held out \textbf{generation tasks} conveying different levels of zero-shot generalization ability: 1) \textbf{weak generalization} set with a random selection of generation tasks with potential overlap of categories with training tasks  and 2) \textbf{strong generalization} set (or strict zero-shot conditions) using summarization and title generation tasks with no overlap in categories from the training tasks. We further investigate the task sets under difficulty levels of \textit{easy}, \textit{medium}, and \textit{hard} based on their baseline ROUGE scores. 
    
     The main conclusion from our study is that under strict zero-shot conditions, meta-learning with instructions significantly improves the performance. The improvements become more significant for the strong generalization task set and when the task difficulty level is hard (i.e. tasks where the LM struggles to generate correct outputs in zero-shot setting). Moreover, meta-learning increases the effectiveness of instructions under all conditions. While both MAML and HNet models show improvements over the baselines, HNet (along with its MAML extension) by explicitly enforcing the use of instructions through task specific conditioning of parameters, results in larger gains. In summary, the main contributions of the paper are two-fold. First, we adapt meta-learning approaches to MTIL. Second, we study their efficacy and show significant improvements under strict zero-shot conditions.


     \section{Related Work}
  




\textbf{Learning from instructions}: An extension of the basic prompt-based in-context learning is appending task specific instructions with prompts. Several recent works which include FLAN~\cite{wei2022finetuned}, T0~\cite{sanh2022multitask}  and~\cite{Reif2021ARF}, train a large LM in a multi-task setting with instructions. 
\textit{InstructGPT}~\cite{Ouyang2022TrainingLM} takes slightly different approach by training the GPT3 model ~\cite{GPT3} with human annotated dataset of demonstrations of desired user intents and use reinforcement learning to improve the model to follow such instructions. Yet another direction called pattern-exploiting training (PET)~\cite{Schick2021ExploitingCF, DBLP:conf/naacl/SchickS21} combines the idea of formulating instructions as cloze questions and show that even small LMs can be good few-shot learners and work with language generation.

\textbf{Meta-learning for language generation}: 
Meta learning has been applied in several language generation settings such as~\cite{MAMLPreTrained} to induce persona in a chatbot,~\cite{Mi2019MetaLearningFL}  for task oriented dialog systems,~\cite{Gu2018MetaLearningFL} for low resource machine translation, and ~\cite{Chen2021MetaTransferLF} for abstractive summarization in a low-resource transfer learning but do not use instructions for zero-shot transfer. Our MTIL scenario is closely related to MetaICL~\cite{Min2021MetaICLLT} which applies multi-task learning in-context in a K-shot setting for classification tasks, but differs in that it is a k-shot in-context scenario and does not use instructions or meta-learning optimization. While these works are related, to the best of our knowledge, meta-learning has not been used to generalize to unseen generation tasks in zero shot settings using instructions and thus the paper provides several novel insights and approaches. 

\textbf{Hyper-Networks (HNet) in NLP applications}:~\cite{karimi-mahabadi-etal-2021-parameter} use HNet to train LMs in a multi-task setting with adapters and~\cite{Oswald2020Continual} propose a continual learning framework with HNets conditioned on unique task IDs to reduce catastrophic forgetting. HNets have been used for input conditioning a decoder in~\cite{hypercoders} which produces a unique decoder for each input, and thus is similar to our approach. However these the approaches are not strictly applicable in our zero-shot scenario or in general NLP tasks with task descriptions in natural language.

\textbf{Language model editing}: Our HNet based approach is based on the architecture in~\cite{DeCao2021EditingFK} which uses it to edit factual knowledge in LMs. While the architecture is similar, we use the HNet to encode task specific instructions and is intended for controlling task-level LM behavior unlike the micro-behavior targeted in~\cite{DeCao2021EditingFK}.  Similar to ours and~\cite{DeCao2021EditingFK}, Bayesian hyper networks~\cite{krueger2018bayesian} linearizes the number of parameters for predictions by constraining the HNet outputs to scale and shift parameters.~\cite{Sinitsin2020Editable, mitchell2022fast} propose Meta Learning approaches for editing errors in a neural network but is not directly applicable for MTIL in a zero-shot setting. 

\textbf{MTIL}: Finally, the work most closely related to this paper is the T\textit{k}-Instruct model from~\cite{WangMishra2022NaturalIB} which fine tunes a T5 model~\cite{Raffel2020ExploringTL} with instructions, which we use as the baseline. We use the same dataset and training settings as  T\textit{k}-Instruct but instead use the pretrained BART model~\cite{Lewis2020BARTDS} as it is task agnostic compared to T5 (T5 may not represent a true zero-shot setting). In addition, we enhance this model with meta-learning and consider significantly different training, evaluation, and model settings to test zero-shot generalization resulting in unique contributions and conclusions orthogonal to the findings in~\cite{WangMishra2022NaturalIB}. 

    \section{Problem Setup}

    In this section we briefly outline our problem settings and baselines used in this paper. 
    
    \subsection{Natural Instructions V2 Dataset}
    We use the Natural Instructions V2 (NIV2) dataset~\cite{WangMishra2022NaturalIB}\footnote{https://instructions.apps.allenai.org/} to investigate meta-learning approaches for instructional learning. The NIV2 is a meta-dataset with over 1600 tasks. 
    
    In NIV2, each task contains instructions and multiple training instances with input and output. The instructions consist of: 1) \textbf{Categories} (classification, summarization etc.), 2) \textbf{Short description} (a short sentence about the task), 3) \textbf{Long description} (a detailed description of the task similar to annotation guidelines), 4) \textbf{Positive examples} (inputs with correct outputs), 5) \textbf{Negative examples} (inputs with \textit{incorrect} outputs), and 6) \textbf{Explanations} for the positive or negative examples.

    ~\cite{WangMishra2022NaturalIB} train a pretrained T5 language model~\cite{Raffel2020ExploringTL} on input-output pairs with instructions (T\textit{k}-Instruct) appended before the input in a multi-task setting. During testing, held out unseen tasks are predicted by appending similar instructions to the test input.~\cite{WangMishra2022NaturalIB} provide detailed ablations and baseline comparisons with related models showing the impact of instructions. Following the results there, we only use the task descriptions and positive examples in this study as negative examples and explanations were not shown to have any positive contributions. 
    
    \subsection{Baseline Model with Standard Training}
    Based on results in~\cite{WangMishra2022NaturalIB} where T\textit{k}-Instruct was shown to comfortably beat much larger T5, GPT3, InstructGPT3, and T0 models, we use the T\textit{k}-Instruct setting as our baseline, i.e. we train a pre-trained encoder-decoder LM on multiple tasks with instructions. We also explored appending the instructions before the decoder sequence but did not find any improvements. However, we did observe that by pre-pending a special prefix to the decoder (we use "[Output]:") improves the overall prediction performance. We refer to this model as the \textbf{standard training} model.  
    
    For our base LM, we use the pretrained BART model~\cite{Lewis2020BARTDS} as it is task agnostic compared to 
    T5\footnote{Publicly available T5 models are pre-trained on a multi-task mixture of unsupervised and supervised tasks.} and thus represents a stronger zero-shot setting. Interested readers should refer to the~\cite{WangMishra2022NaturalIB} paper for detailed ablations specific to the NIV2 dataset and the T5 model.
    
    \subsection{Evaluation Settings}
    We focus specifically on the zero-shot generalization on generation tasks. 
    While the general settings remain similar to ~\cite{WangMishra2022NaturalIB} we consider some specific settings to illustrate the generalization capabilities of models to different tasks. 
    
    For training, we use two sets of tasks 1)  All EN tasks in the NIV2 dataset and 2) Generation tasks. For evaluation, we consider two sets of generation tasks with different zero-shot levels : 1) weak generalization set using a random set of generation tasks with potential similarity to the training tasks and 2) strong generalization set using tasks from summarization and title generation categories with no overlap with the training tasks. The list of evaluation tasks with short descriptions are provided in the appendix in Figures~\ref{fig:WeakGenerationSet} and~\ref{fig:StrongGenerationSet}.
    
    We further divide the evaluation tasks into difficulty levels of "easy", "medium" and "hard" based on the ROUGE scores from the baseline model (low scores indicate out-of distribution and difficult tasks) to see to what extent meta-learning helps in improving performance of the out-of-distribution tasks.

\section{Meta-Learning with Instructions}\label{sec:model_description}

    Training on a large number of diverse tasks and testing on unseen tasks lend itself to the paradigm of learning-to-learn or meta-learning, which has been successfully applied for generalizing in both zero- and few- shot scenarios. Task meta-data in the form of instructions can also provide discriminative information about the task process in addition to the surface patterns of the input and output strings. We investigate whether meta-learning can aid such learning and adapt three approaches to MTIL.


 \subsection{Standard Training + MAML} 
    We adapt Model Agnostic Meta Learning (MAML) ~\cite{Finn2017ModelAgnosticMF} to instructional learning of LMs as a way to generalize to unseen tasks by training on large number of diverse tasks. 
    
    
    The standard training with MAML is described in Algorithm~\ref{alg:MAML} in the appendix. At any training iteration, we sample two different sets of $k$ tasks for MAML meta-train and meta-test steps. We uniformly sample across tasks to maximize the diversity of tasks in each batch. The data format is same as the standard training. Since we test zero-shot conditions, we do not have any test time optimization typically employed in MAML. 
    
    
    \begin{figure}
        	\centering
        	\includegraphics[scale=0.280,trim={0 170 0 80},clip]{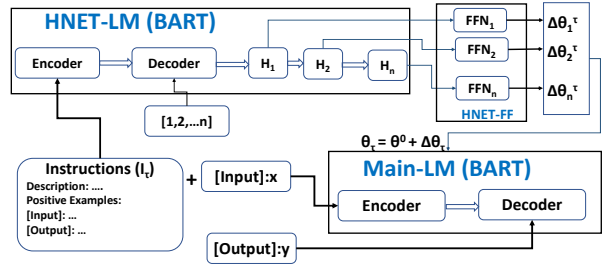}
        	\caption{Encoding instructions using a Hyper-network}
        	\label{fig:HNETDiagram}
        \end{figure}
    
    \subsection{Standard Training + HNET} 
         Both standard and MAML training do not explicitly enforce the use of instructions during decoding. The model can thus minimize the loss simply by ignoring the instruction part of the encoder by attending to the input and output texts. This can lead to sub-optimal use of the instructions. 
         
         We propose a hyper-network (HNet) based architecture ~\cite{Ha2017HyperNetworks} to produce task specific model parameters conditioned on the instructions. The HNet architecture consists of an auxiliary LM (HNet-LM) along with the Main-LM. The HNet-LM produces task specific parameters of the main LM by conditioning on instructions.
         
         In particular, we adapt a specific type of HNet architecture from ~\cite{DeCao2021EditingFK} 
         which predicts the \textit{delta-parameters} of the Main-LM, which are then added to the Main-LM parameters to produce the task specific LM. This preserves the parameters in the Main-LM utilizing the shared generation capability of the LM while specializing in task specific behavior.  However, there are some specific differences based on our requirements for instructional training, which are described next. 
         
         \subsubsection{The HNet Language Model (HNet-LM)}
         
         Since the input to the HNet model is text, we use a pretrained encoder-decoder LM (BART in this paper) to encode the instructions\footnote{In contrast ~\cite{DeCao2021EditingFK} used an untrained LSTM.} and use the decoder's hidden states for conditioning the layer specific parameters of the Main-LM.
         
         In ~\cite{DeCao2021EditingFK} the last hidden state from an LSTM is used for conditioning the parameters of the main model. To increase the effective bandwidth of the HNet while keeping the number of parameters same, we use the last $N$ hidden states (for $N$ layers of the main LM). This simple trick allows the model to independently attend to and condition each layer on the input instructions while still keeping the model parameters the same. 
        
        The HNet-LM takes instruction $I_{\tau}$ for a task $\tau$ and sequence of decoder indexes $d_n$ 
        as input and produces $N$ hidden states  $h_{\tau}(n)$. The decoder index sequence we use is simply $1,...,n$. Decoder indexes provide different inputs to the decoder to influence the generation of distinct parameters for each layer of the main LM. This is not strictly required as the position embeddings can in principle drive the input/output of the HNet to produce different parameters for each target layer. However, we found that adding different decoder indexes improves the performance as it provides additional differentiation to the decoder.
        
        The HNet-LM can produce the parameters of all or a subset of layers of the Main-LM. We experimented with three settings: encoder, decoder, and both. The best metrics are obtained when the HNet is used to generate parameters of the decoder of the Main-LM, which is what we use in reporting the results.
          
          \subsubsection{HNet-FF Projections}
          Next, the output hidden states from the HNet-LM decoder are projected to the Main-LM parameter space using the HNet Feed Forward (HNet-FF) layer consisting of one dense layer with $Tanh$ activation. Let $\psi$ denote the parameters of the HNet-LM, $\pi_n$ the parameters of the $N$ HNet-FF layers, and $d_n$ an input index to the decoder.

          The $h_{\tau}(n)$ is then projected using the FF network to five parameter vectors $\alpha_n, \beta_n, \gamma_n, \delta_n, \eta_n$ as given in Equation~\ref{eq:hnet_ff}.  Finally the delta parameters $\Delta \theta_{\tau}$ of the main LM layers are generated using equation~\ref{eq:hnet_transforms}. Here $\sigma$ denotes the sigmoid function and $\sigma'$ is the softmax. The projection process is equivalent to  ~\cite{DeCao2021EditingFK} but slightly differs in that we do not use the gradients of the Main-LM parameters  as we target a zero-shot scenario and during test time, the target labels are not available. The model is illustrated in Figure~\ref{fig:HNETDiagram} and Algorithm~\ref{alg:HNET} in the appendix. 
        
         {\small
                \begin{gather}
                    h_{\tau}(n) = HNetLM(\psi;I_{\tau}, d_n) \\
                    \alpha_n, \beta_n, \gamma_n, \delta_n, \eta_n \gets HNetFF_n(\pi_n; h_{\tau}(n)) \label{eq:hnet_ff}\\
                    \Delta\theta_{\tau}  = \sigma(\eta) \cdot (\sigma'(\alpha)\gamma^T + \sigma'(\beta)\delta^T) \label{eq:hnet_transforms}\\
                    \theta_{\tau} = \Theta_{0} + \Delta\theta_{\tau} 
                    \label{eq:hnet_delta_params}
                \end{gather}
            }
    \subsubsection{Alternating Training Schedule}
       The HNet-LM along with the Main-LM can be jointly trained end-to-end. We tested with different configurations such as partially or fully freezing the Main-LM, but best metrics were achieved with the Main-LM fully trained along with the HNet. However, with this comes convergence issues and significantly increased training cost. First, the training can be unstable when both the HNet and Main-LMs are updated, requiring low learning rates. Second, joint training requires twice as much memory and computation.

       We address both issues by using an alternating training schedule for the HNet and Main LMs, where we freeze one of the LMs for a few steps while updating the other and vice-versa. Thus, in the backward step only one of the LM's parameters are updated, which leads to lower memory requirements, stable loss convergence, and better metrics. Moreover, during the HNet training phase (i.e., when the Main-LM is fixed), the model is forced to update the HNet parameters using the instructions as input thus allowing itself to learn instruction-specific patterns. The HNet based model training is described in Algorithm~\ref{alg:HNET} in the appendix.

\subsection{Standard Training + HNet + MAML}
    Since the HNet model comprising of the HNet-LM, HNet-FF, and Main-LM is end-to-end differentiable, we can employ MAML training loop for this combined network. Here, inside the HNet loop, we employ the  alternate training schedules in the MAML inner loop, while the outer training loop remains the same.  The HNet-MAML training is described in Algorithm~\ref{alg:HNET_MAML} in the appendix.

    \section{Experiments}

\subsection{Experimental Setup}
    We use the NIV2 dataset in all our experiments in a multi-task setting similar to prior work ~\cite{WangMishra2022NaturalIB} for training LMs to interpret instructions. I.e. we train on a large number of tasks  with instructions and analyze  the zero-shot generalization capability on held-out evaluation tasks. 
    
    We first split the instances in NIV2 dataset for each each task into training, validation, and evaluation sets consisting of 80\%, 10\%, and 10\% split respectively. We filter out any non-English tasks for this study, which leads to around 1000 tasks. We consider two sets of training and evaluation tasks which are intended to better understand the generalization capability of LM with instructions. We first create two tasks sets as follows: 

    
    
    \textbf{1) All-tasks set} which has all EN tasks from NIV2 dataset consisting of around 1000 tasks. 
    
    \textbf{2) Generation-tasks set} which consists of  language generation tasks by filtering out task categories such as classification, sequence tagging, multiple choice QA, etc. This consists of around 400 tasks.
    
    Next we randomly split (90\%/10\%) the Generation task set into training and evaluation tasks. Additionally the tasks from the held out strong generalization set (described next) are also removed from the training sets. After filtering, we have 916 tasks in the All-tasks training set and 306 tasks in the Generation-task training set. 
    
    
    For evaluation, we consider two language generation task sets which differ in the level of dissimilarity from the training tasks and thus test different levels of generalization ability of the trained LMs. 
    
    \textbf{Weak generalization evaluation set (81 tasks)}: This set (also used as a validation set) comprises of 10\% of tasks randomly split from the Generation task set. Since it is a random split, this set has overlap in the categories with the training set.
    
    \textbf{Strong generalization evaluation set (33 tasks)}: This set consists of tasks with no overlap in categories with either the training or the validation sets. We select text compression categories of Summarization and Title Generation tasks as they require complex understanding of language components, and are sufficiently distinct from training tasks (majority of which are Q/A generation, text manipulation etc.).

    While both the evaluations sets are held-out during training, second set tests a stricter level of zero-shot generalizability to unseen tasks. 
    
    \textbf{Instruction components}: We use four configurations of the instruction components appended to the input of train instances: \textbf{1) None} uses the short task description, \textbf{2) Desc} uses the long task description. \textbf{3) PosEx} adds one positive example of input/output pairs, and \textbf{4) Desc + PosEx} includes both the task description and one positive example. 
    
    \textbf{Training parameters}: 
    For training, we randomly select a maximum of 100 instances from each task, and  train for around 20 epochs\footnote{Similar to ~\cite{WangMishra2022NaturalIB}, we see that more instances or epochs can overfit on the training data and lead to poor zero-shot generalization.}. We use the pre-trained BART-base model for most of our experiments and analysis (BART-large numbers are also briefly reported). We use maximum context lengths of 1024 (encoder) and 128 (decoder) tokens. We do not truncate the sequences but instead filter out sequences which do not fit in the context length. We make sure that all models and configurations get exposed and evaluated on the same data by pre-filtering out instances for the longest input configuration (Desc + PosEx). All models are trained on a single NVIDIA-V100 GPU with 32GB memory, and takes around 24-72 hours (depending on the model complexity and data size) to train. More training details are provided in the appendix.
    
    For evaluation, we follow the NIV2 settings where a maximum of 100 instances per task was used for evaluation to reduce overhead for large scale experimentation. This leads to around 3000 instances for the smaller strong generalization set, and around 5000 for the larger weak generalization set with multiple references per instances. We generate predictions using greedy decoding and compute ROUGE-2F and ROUGE-L (both follow similar trends with minor differences). However, we found that ROUGE-2F was slightly more robust with less variance in the zero-shot scenario, when in the initial stages of training, the model frequently copies the input prompt/instruction to output giving unnaturally high ROUGE-L scores. ROUGE-2F was less susceptible to such a scenario and thus we use ROUGE-2F metrics for our analysis (ROUGE-L is reported in the appendix). We compute both the overall metrics as well as the task level metrics and analyze them in detail. 
    
    For implementation, we adapted code from several open source packages: HuggingFace Transformers ~\cite{wolf-etal-2020-transformers}, learn2learn~\cite{learn2learn}, Higher~\cite{grefenstette2019generalized}, and KnowledgeEditor~\cite{DeCao2021EditingFK}\footnote{https://github.com/nicola-decao/KnowledgeEditor}. Our adaptations and complete training and evaluation scripts will be open sourced for further research.

    \subsection{Summary of Results}
    We first provide a short summary of conclusions from our study before diving into the details. 
    
        \textbf{Task descriptions}: Descriptions improve performance when used by itself. When used with positive examples, it improves the performance, for the strong generalization set. For the weak set, positive examples are sufficient. 
        
        \textbf{Training sets}: The all-tasks set has better performance even though the generation-set is more similar to the evaluation tasks. I.e. it is better to train on a larger number of diverse tasks for better zero-shot generalization. 
        
        \textbf{Evaluation sets}: For the strong generalization set, instructions and meta-learning improve the performance. For the weak set, meta-learning improves performance only when using task descriptions without demonstrations or with the smaller training set.
        
        \textbf{Model performance}: Both MAML and HNet models improve performance but HNet with task specific encoding is better. HNet-MAML improves the performance by almost 30\% overall in the strong generation set, showing the effect of twin enhancements. 
        
        \textbf{Task difficulty level}: Meta-learning significantly improves performance (by almost 100\%) for "hard" tasks, with HNet-MAML having the best performance. Meta-learning also significantly increases the impact of using instructions: impact increases from 250\% for standard to 1500\% using HNet-MAML.

    \begin{figure}
    	\centering
    	\includegraphics[scale=0.41,trim={0 160 0 80},clip]{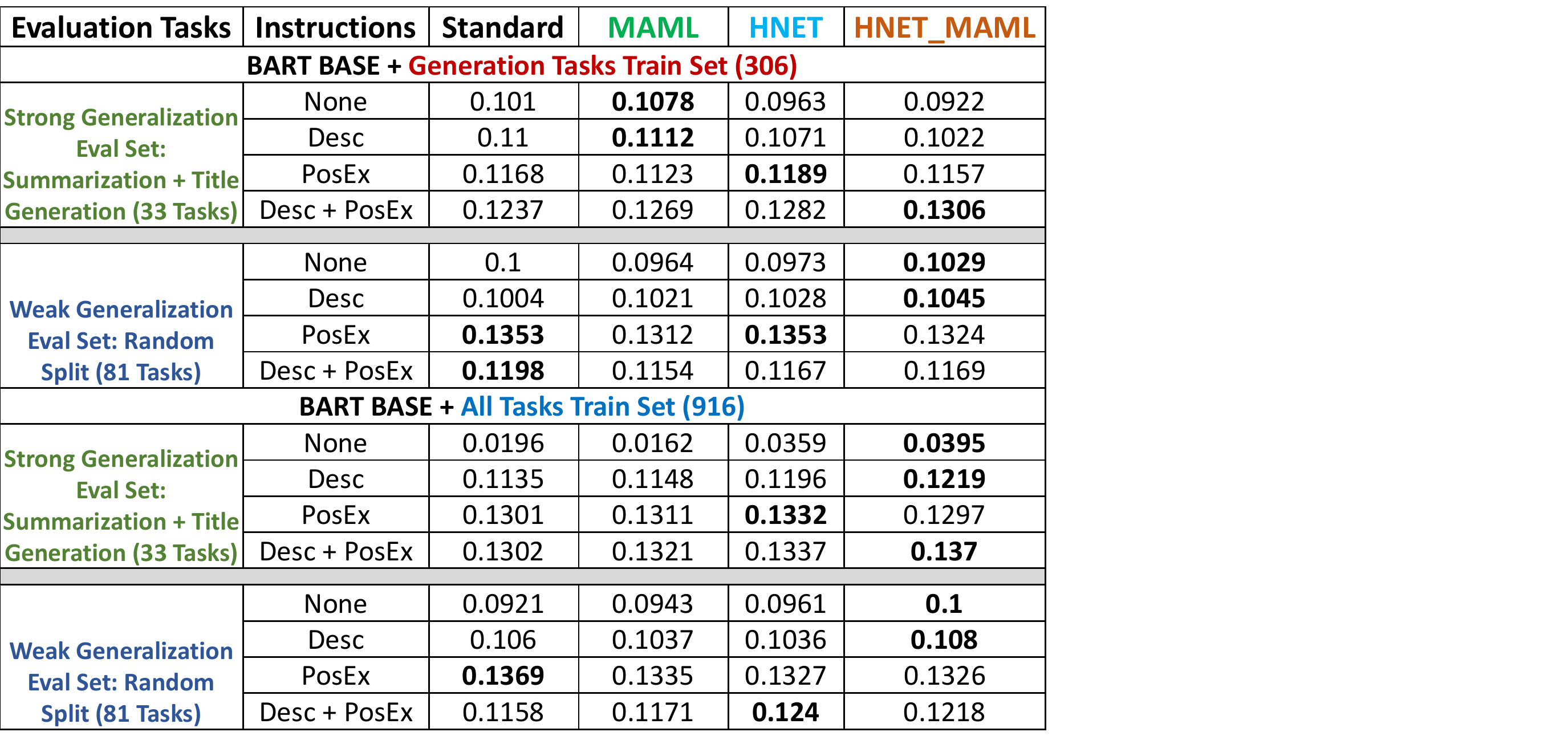}
    	\caption{Metrics (ROUGE-2F) with different models and instruction fields, on the two training and evaluation sets. Instructions improve the performance specially for the strong generation set. Best metrics are obtained with HNet-MAML using using Desc + PosEx.}
    	\label{fig:OverallAblationsTable}
    \end{figure}

    \subsection{Performance of Different Models}
    
    Baseline metrics on the standard training with ablations on instruction components, and a short ablation study comparing the different models are presented in the appendix in Figures~\ref{fig:ShortAblationsTable} and~\ref{fig:StandardAblationsTable}.
    
    Here, we compare the overall performance of the standard model with the meta-learning approaches in Figure~\ref{fig:OverallAblationsTable} which reports the overall ROUGE scores for the different evaluation sets. 
    
    \textbf{Role of instructions}: Instructions improve metrics for the strong generalization set across all the models and for both the training sets. As the instructions get more detailed (None through Desc+PosEx) we see improved performance with the more complex models. When no instructions are used (None), MAML performs the best. When entire instruction is used (PosEx+Desc) HNet-MAML has the best performance. The results illustrate that instructions are more effective with meta-learning, particularly for the strong generalization set. However, for the weak generalization set, the surface patterns of the positive examples are sufficient and task descriptions do not improve the generalization capability. In addition, the standard training achieves the best performance for the weak set showing that meta-learning is not as effective for weak generalization conditions. 
    
    \textbf{Training sets}: Overall, the performance is better with the larger training set but mostly for the strong generalization set. For the weaker set, the smaller training set of just 306 tasks is competitive with the larger set (achieved with just the positive examples and standard training), showing that task similarity in train and evaluation does matter. However, under stricter zero-shot conditions it is better to have a more diverse and larger set of training tasks.

    \textbf{Strong generalization eval set}: We see improved performance of MAML and HNet variants for the strong generalization set for both sets of training tasks. When instructions are most detailed (Desc + PosEx), HNet-MAML has the best performance with both the training sets. The results illustrate the effectiveness of creating task specific model parameters through the HNet especially under strict zero shot settings. Moreover, HNet-MAML has the best performance showing that the twin optimizations utilize instructions and generalize better to out-of-distribution tasks. 
    
    \textbf{Weak generalization eval set}: For the weak set (when the model has seen
    similar tasks in training), task descriptions are not useful and best metrics are achieved using just the positive examples across all models. It is also interesting to note that the smaller train set has comparable metrics with the larger set, showing that the set can learn mostly from the set of tasks in the generation train set. Moreover, results here show that standard training from~\cite{WangMishra2022NaturalIB} is a strong baseline and thus further distinguishes the performance of meta-learning in utilizing instructions under strict zero-shot conditions.  
    

    \begin{figure}
    	\centering
    	\includegraphics[scale=0.65,trim={0 270 480 80},clip]{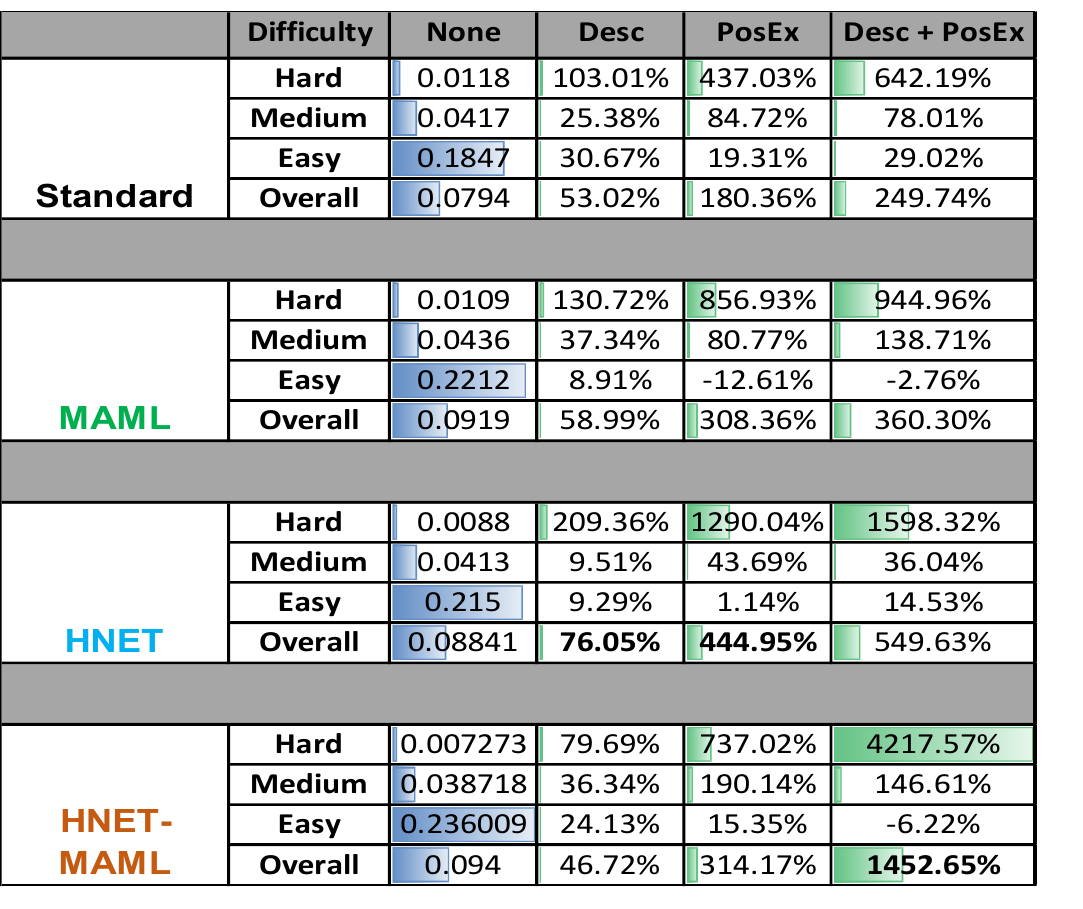}
    	\caption{The \% improvements with different instruction components. Tasks are split into easy/medium/hard based on the ROUGE-2F scores from standard training with the None setting. Instructions help the hard task across all models with best results using HNet-MAML.}
    	\label{fig:CompareTaskLevelInstructions}
    \end{figure}

    \subsection{Effect of Instructions on Tasks with Different Difficulty Levels}

    Next, we analyze how the models utilize the different instruction components by breaking down the performance to different difficulty levels in Figure~\ref{fig:CompareTaskLevelInstructions}. We compute the ROUGE scores for each task separately, and deem tasks whose metrics from the baseline (standard training) are low as \textit{hard} and ones with high as \textit{easy}. Then we sort the tasks by the scores and divide them into three difficulty groups: "easy",  "medium" and "hard". For the strong generalization set of summarization and title generation tasks (33 tasks), we have 11 tasks while for the weak generalization set (81 tasks), we have 27 tasks per group. We report the \% change in the metrics with different instruction components for each model with the "None" setting as the baseline in Figure~\ref{fig:CompareTaskLevelInstructions}. 
    

    
    \textbf{Instructions help hard tasks}: Figure~\ref{fig:CompareTaskLevelInstructions} shows that instructions have the biggest impact on the hard tasks and when both positive examples and descriptions are used. For example, we get an improvement of 642\% with standard training and 4000\% for Hnet-MAML using instructions (compared to the None setting). For easy tasks, instructions can even lead to regression in metrics. For example, we see a -6.22\% regression with HNet-MAML model. 
    
      \begin{figure}
    	\centering
    	\includegraphics[scale=0.45,trim={0 170 0 80},clip]{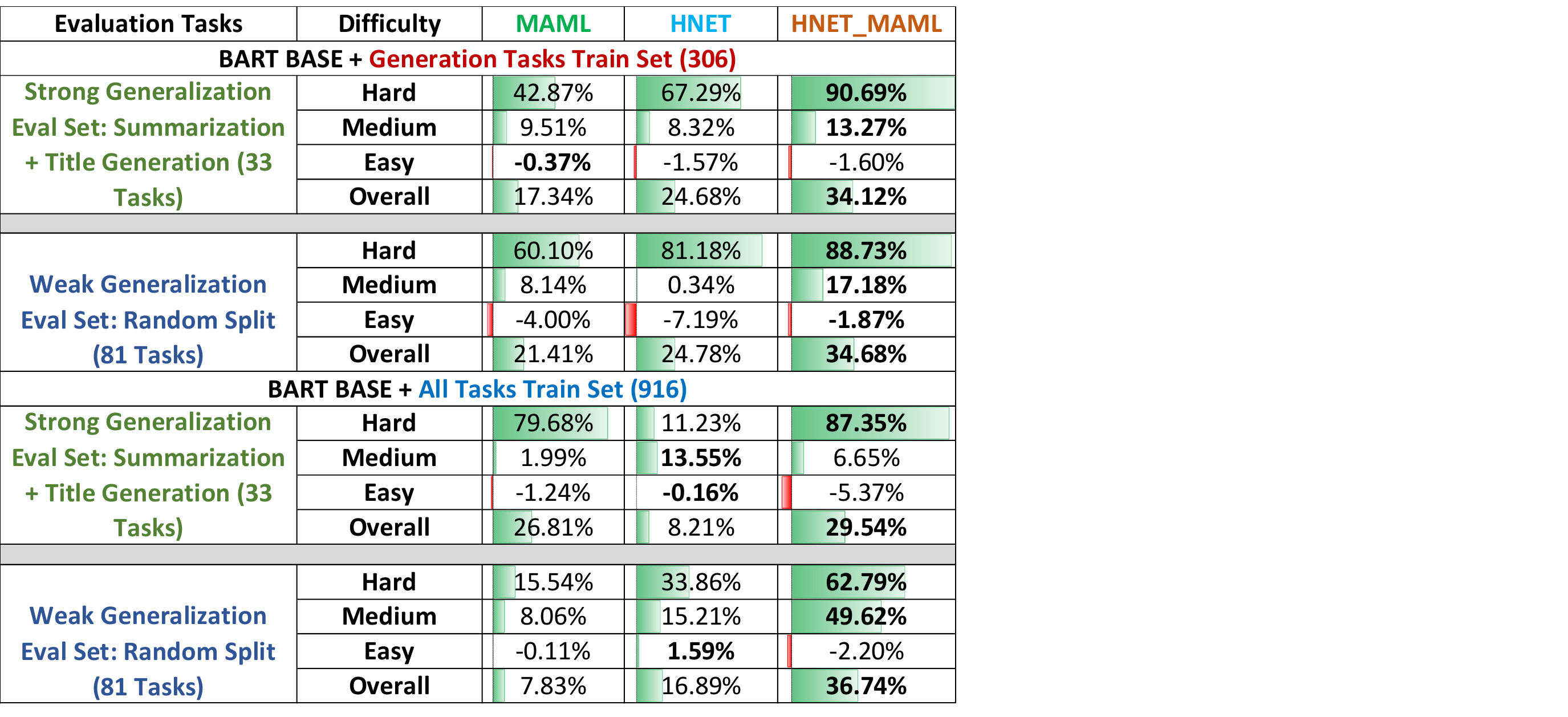}
    	\caption{\% differences from standard training with other models for the two training and evaluation sets. 
    	\textbf{HNet-MAML} and \textbf{hard tasks} have the largest improvements across the different train and eval sets. }
    	\label{fig:CompareTaskLevelDifficultyGenerationTasks}
    \end{figure}
    
    \textbf{Meta-learning increases the effectiveness of instructions}: Figure~\ref{fig:CompareTaskLevelInstructions} also shows that meta-learning can significantly boost performance. Overall, HNet-MAML is able to increase the performance by 1452\% using instructions over the None setting compared to around 250\% for standard, 360\% for MAML, and 549\% for HNet. Thus, while MAML is able to utilize the instructions better than standard training, HNet by explicitly conditioning the parameters on instructions can improve it further for out-of-distribution tasks where the additional information from instructions are most useful. In addition, we see that the twin meta-learning optimizations with HNet and MAML maximize the utilization of instructions. 
    

    \subsection{Detailed task level analysis of models}
    
    Next we analyze the metrics across the two training and evaluation sets for the different models. Here we employ the full instruction (Desc+PosEx) and use the metrics from standard training. We divide the tasks into the three difficulty levels of easy/medium/hard and report the \% changes with other models in Figure~\ref{fig:CompareTaskLevelDifficultyGenerationTasks}. 
    
    It is interesting to contrast Figure~\ref{fig:CompareTaskLevelDifficultyGenerationTasks} with the overall metrics reported in Figure~\ref{fig:OverallAblationsTable}. While the overall metrics show small ROUGE differences, breaking down the metrics at the task difficulty levels illustrate significant differences. This is due to the large range of ROUGE scores across tasks, which can normalize the overall differences. 

    \textbf{Train sets}: In Figure~\ref{fig:CompareTaskLevelDifficultyGenerationTasks}, we see that relative performance improvement with meta-learning approaches are similar for the two training sets across the models even if the absolute numbers differ. HNet-MAML has the best performance across the two train sets with an overall improvement of around 30\% compared to standard training.

    \textbf{Eval sets}: For the two eval sets, when broken down into task difficulty levels, we surprisingly see that meta-learning models are better for both the weak and the strong sets. This is because even in the weak set, there are hard tasks for which the meta-learning models improve the generalization. Moreover, the hard tasks get the highest improvements using HNet-MAML and shows the impact of twin optimization on out-of-distribution tasks inside both weak and strong  generalization sets.   

     \begin{figure}
    	\centering
    	\includegraphics[scale=0.74,trim={0 170 500 80},clip]{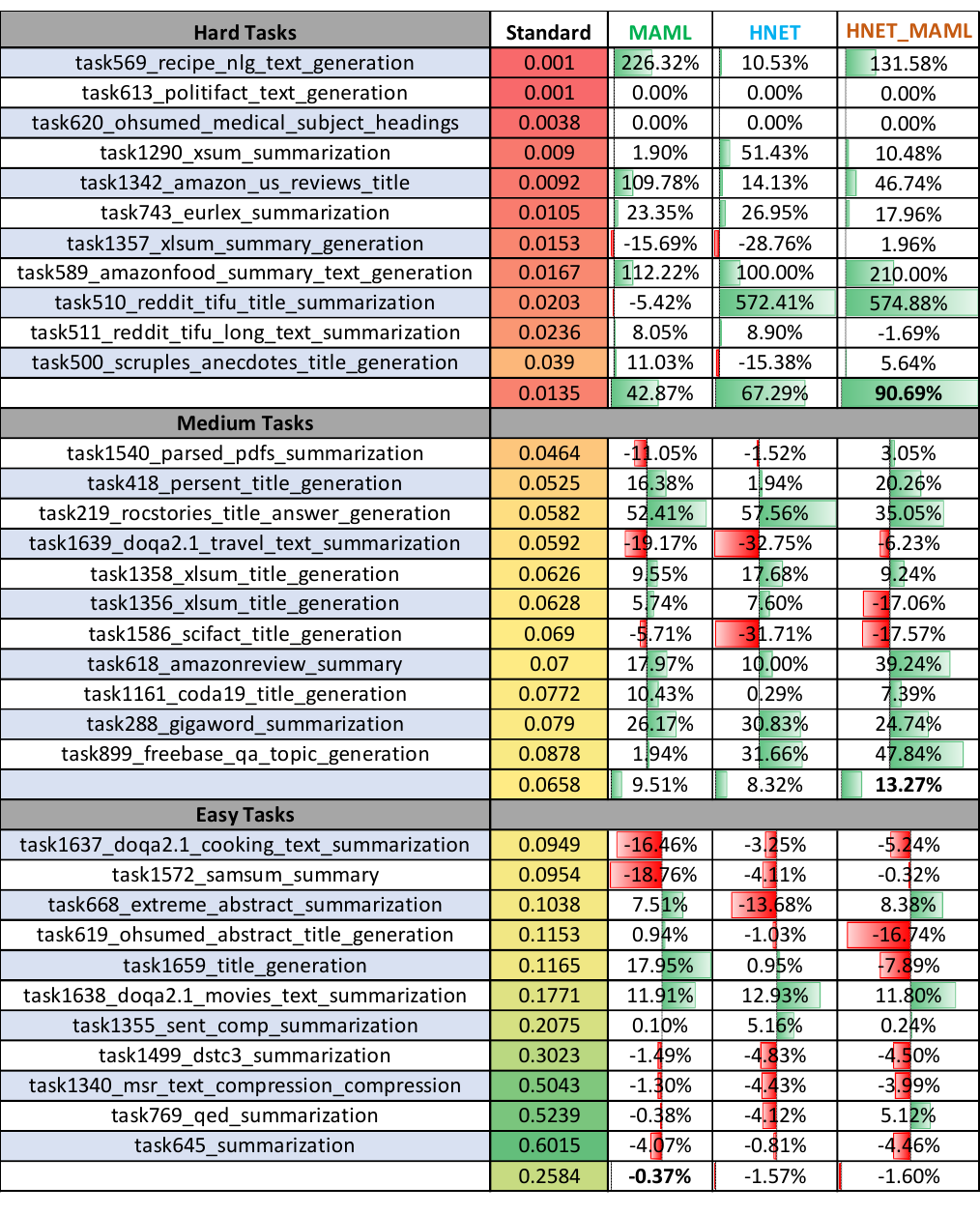}
    	\caption{\% differences listed for individual tasks divided into easy/medium/hard difficulty levels. Results show that MAML, HNet and HNet-MAML models have significant improvements for the difficult tasks)
    	}
    	\label{fig:DetailedTaskLevelDifficultyGenerationTasks}
    \end{figure}

    \textbf{Per-task metrics}: We report the per-task metrics for the three groups in Figure~\ref{fig:DetailedTaskLevelDifficultyGenerationTasks} for the strong generalization set. The standard model does really poorly in some of the hard tasks. Essentially these are tasks where the patterns are significantly different from the training tasks and the model is unable to generalize to the new instruction patterns. This is where meta-learning and in particular the twin optimization of MAML and HNet significantly improve the scores. This improvement can be the key factor whether zero-shot based predictions can be used practically and can result in big difference in how users perceive the model quality. 
    
\begin{figure}
    	\centering
    	\includegraphics[scale=0.31,trim={0 250 0 80},clip]{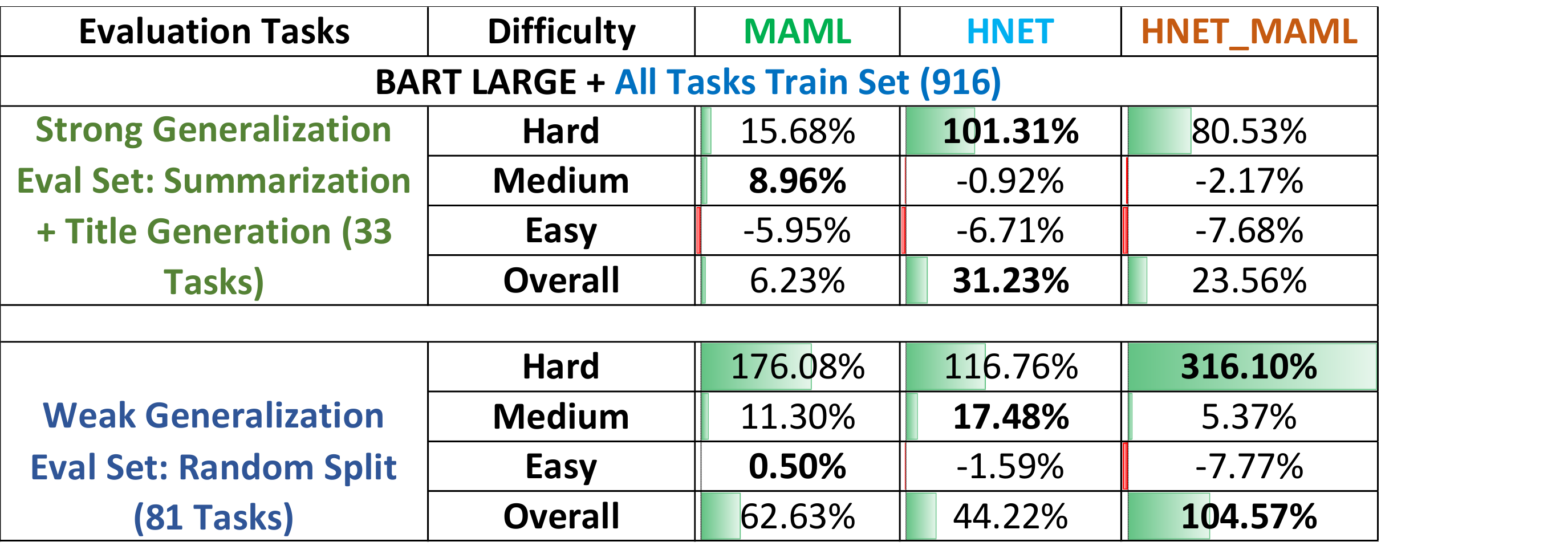}
    	\caption{\% differences from standard training with other models with \textbf{BART-Large}. \textbf{HNet} has the best performance for the strong generalization set.}
    	\label{fig:CompareTaskLevelDifficultyBartLarge}
    \end{figure}

    \textbf{BART-Large}: We also report the task level metrics for BART-Large in Figure~\ref{fig:CompareTaskLevelDifficultyBartLarge}. Meta-learning has a higher impact with BART-large compared to BART-Base. However, the performance for HNet and HNet-MAML is mixed. HNet-MAML with BART-large is difficult to train on a single GPU due to high memory requirements (requiring small batch sizes), which might have reduced its effectiveness (See appendix on the training parameters for different models and further discussion). 
    
    \section{Conclusions}
In this paper we investigate whether  meta-learning applied to multi-task \textit{instructional learning} (MTIL) can boost the generalizability of LMs to unseen tasks in a zero-shot setting. Specifically, we evaluate MTIL in three directions with MAML, HNet and HNet-MAML. To test the generalization ability, we consider two sets of training and evaluation task sets and through extensive experiments on the NIV2 dataset, show that meta-learning can significantly boost the performance by increasing the effectiveness of instructions particularly under strict zero shot conditions and for "hard" tasks. 

While the models perform relatively well under zero-shot conditions, the performance is far from fully supervised models. It remains to be seen at what point we can match fully supervised models (for example using a k-shot setting). In addition, the impact of HNet-MAML on the BART-Large model was lower. It will be interesting to see how meta-learning scales with model sizes and whether the additional bandwidth from larger models can negate the impact of meta-learning in encoding and utilizing instructions.  This is subject of future work.


\section{Limitations}
There are several limitations of the proposed meta-learning based approaches in its present form. 

\begin{itemize}
    \item Computation and memory overhead: Meta-learning approaches have higher resource requirements which can limit the usage specially for larger models. For example with BART-large, the HNet-MAML model on a single GPU is inefficient to train since we have to use small batch sizes which leads to lower performance. 
    \item Regressions with easy tasks: We see some regression in metrics for the easy tasks. Further analysis and research is needed to understand the factors and improve the models such that model enhancements are uniform across tasks. 
    \item Hyper-parameter tuning: Meta learning models have more hyper-parameters and thus might be more difficult to tune than the standard training approach. 
    \item Overall zero-shot performance: The zero-shot performance even with the best meta-learning approaches is quite far from state-of-the-art results. It will be interesting to see at what point (e.g. with $k$-shot learning) the performance can match a fully supervised model. 
\end{itemize}
    \bibliographystyle{acl_natbib}
    \bibliography{main_camera_ready}
    \begin{appendix}
      \section{Model Details}
    
    \subsection{Standard Training with MAML}
        The training loop for MAML with instructions is given in Algorithm~\ref{alg:MAML}. At each step, we sample two sets of tasks for the MAML fast-adaptation and meta-update steps. To increase the generalization of the process, we force tasks to be unique across the two sets. This forces the use of different sets of instructions for the two MAML steps. We also sample tasks uniformly to maximize the diversity of tasks the model gets exposed to during training. These factors improve the performance over a proportionate sampling method.
        
        For MAML training, we consider the following hyper-parameters: inner and outer learning rates, the inner batch sizes, the number of inner loop iterations, and the number of tasks in the loop. We use first-order MAML as it takes significantly more memory to maintain the second order gradients and fit large batches in a single GPU.
        
        We conducted ablations studies to find the best hyper-parameters for MAML. It is important to note that the hyper-parameters are constrained by memory. For example if we increase the batch size, or the number of tasks, we have to reduce the number of inner steps to keep the training memory within the GPU memory limits. Based on our ablations we use 3 inner steps, a batch size of 10, and 2 tasks per MAML step. We use a inner learning rate of $5e-3$ and an outer learning rate of $5e-4$ for all the experiments. For BART-Large we use a batch size of 4. With gradient accumulation steps we can increase the effective batch sizes to up-to around 400 for BART-base and 200 for Bart-Large.   

            \begin{algorithm}
            {\small
                \caption{MAML Loop with Instructions}\label{alg:MAML}
                $MAML(\theta_{MainLM}; I_{\tau}, x, y)$
                \begin{algorithmic}
                    \State Sample K Support and Target tasks $\tau_s, \tau_t$
                    \State $\Theta_0 \gets \theta$
                    \For{$k \in 0:K_{tasks}$}
                        \For {$n \in 0:N_{steps}$}
                            \State $\Theta_{n+1,k} \gets GD\mathcal{L}(\Theta_{n,k},, (I_{\tau_{s}}, x, y)$
                        \EndFor
                    \EndFor
                    \State $\theta = \Theta_{N} - \beta \nabla_{\Theta_0} \sum_{k} \mathcal{L}(\Theta_{N,k} ; (I_{\tau_{t}}, x, y))$
                \end{algorithmic}
            }
        \end{algorithm}
        
        \subsection{Standard Training with HNET}
        
        Standard training with HNet is described in Algorithm~\ref{alg:HNET}. The HNet model consists of HNet-LM, HNet-FFs and the Main-LM. We use the BART model for both the HNet-LM and Main-LM. While potentially any pretrained text encoder could have been used for the HNet-LM, we keep the two LMs the same. This is mostly a practical consideration which allows us to use a single tokenizer to process the text. This also maintains the uniformity of input across models (for examples instructions fed into the HNet-LM  vs. Main-LM) and allows flexibility in the use the tokenized text in either LMs. 
        
        {\small
            \begin{algorithm}
                \caption{HNet with Instructions}\label{alg:HNET}
                $HNET(\theta_{MainLM}, \Phi_{HNetLM};I_{\tau}, x, y)$
                \begin{algorithmic}
                    \State Sample a task $\tau$ and mini-batch from the task
                    \State $\Delta\theta_{\tau} \gets \Phi(I_{\tau})$
                    \State $\theta_{\tau} \gets \theta_{Main} + \Delta\theta_{\tau}$
                    \If{Alternating}
                        \If{$steps\%k = 0$, Freeze $\theta$}
                            \State $\Phi' \gets \mathcal{L}_{GD}(\Phi;\theta_{\tau}(I_{\tau}, x, y))$
                        \Else{Freeze $\Phi$}
                            \State $\theta' \gets  \mathcal{L}_{GD}(\theta; \theta_{\tau}(I_{\tau}, x, y))$
                        \EndIf    
                    \Else 
                        \State $\theta',\Phi' \gets \mathcal{L}_{GD}(\Phi, \theta; \theta_{\tau)}(I_{\tau}, x, y))$
                    \EndIf
                \end{algorithmic}
        \end{algorithm}
        }
                    
    \begin{figure}
    	\begin{center}
        	\includegraphics[scale=0.84,trim={0 300 500 80},clip]{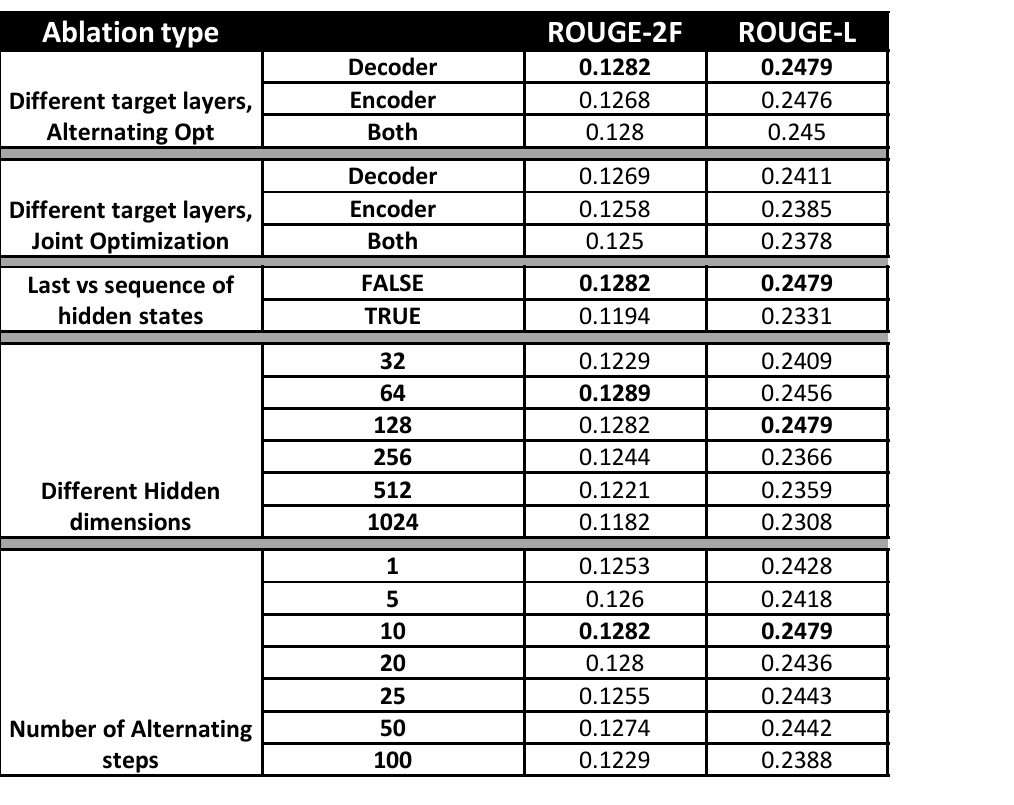}
        \end{center}
        \caption{Ablation studies with HNET model.}
        	\label{fig:HNETAblationsTable}
    \end{figure}
         
       During training, each instance for a given task produces a unique task specific LM. This prevents training in batch-mode when the batch consists of a random set of tasks, and considerably slows down the training. To speed up training, we always create a batch of instances from the same task such that at any step there is a single task specific LM through which we do the forward and backward steps. While this reduces the metrics to an extent, it leads to much faster training of the HNet models.  
        
       We conducted extensive ablations with the HNet model to understand the best hyper-parameters. The ablations are based on the following hyper-parameters and configurations, shown in Figure ~\ref{fig:HNETAblationsTable} and discussed below.
    
        \begin{itemize}
            \item HNet target layers: Since the Main-LM has two transformer stacks in the encoder and decoder, we consider predicting the parameters of encoder, decoder, or both using the HNet. Targeting the decoder has the best performance.

            \item Last vs. sequence of output hidden states from HNet-LM: Since the HNet-LM is also an encoder-decoder model, we compare using last hidden state vs. the sequence of hidden states for projecting to the Main-LM parameter space. We find that sequence of hidden steps has better performance. 
            
            \item HNet hidden dimension: The dimension of the hidden layer for the FF projections for each layer. Our ablations show that the best metrics are achieved with a dimension size of 128.
            
            \item Training Schedule: Joint vs. Alternating. We found that the alternating schedule has better performance.  Having a schedule frequency of 10 steps leads to the best results.
        \end{itemize}
        
        {\small
        \begin{algorithm}
            
                \caption{HNET-MAML with Instructions}\label{alg:HNET_MAML}
                $HNETMAML(\theta_{MainLM}, \Phi_{HNetLM}; I_{\tau}, x, y)$
                \begin{algorithmic}
                    \State $\Theta_0 \gets [\theta_0, \Phi_0]$
                    \State Sample Support and Target tasks $\tau_s, \tau_t$ 
                    \State Sample mini-batches from $\tau_s, \tau_t$
                    \For {$n \in 0:N$}
                        \State $\theta_{n+1},\Phi_{n+1} \gets HNET(\theta_n, \Phi_n;I_{\tau_s}, x, y)$
                    \EndFor
                \end{algorithmic}
                \begin{algorithmic}    
                    \State $\Theta_0 \gets \theta_{N} - \beta \nabla_{\theta_0} HNET(\theta_n, \Phi_n;I_{\tau_t}, x, y)$
                \end{algorithmic}
        \end{algorithm}
        }
        
    \subsection{HNet-MAML model}
    
    The HNet-MAML model is described in Algorithm~\ref{alg:HNET_MAML}. Here, the input model to the MAML loop is the HNet model comprising of the HNet-LM, HNet-FFs and the Main-LM. Within the MAML inner steps, each individual step uses the HNet inner step (i.e. first project the instructions to the Main-LM parameter space and then generate using the Main-LM). Similar to the HNet training, we use the alternating training schedule for the HNet inner loop. The HNet's inner alternating frequency takes into account the number of inner MAML steps, gradient accumulation steps and extra steps due to the fast adaptation and meta-update steps of the outer MAML loop to ensure that  only one of the LMs are updated during the inner and outer MAML loops.

    For BART-base we use an inner batch size of 10, 3 inner steps and 2 tasks per MAML loop. For BART-Large, we use an inner batch size of 2, with 3 inner steps and 2 tasks (one each for MAML train and test steps). As with the HNET model, to enable batch processing, we sample instances in each step from the same task. Using gradient accumulation, with BART-base models we are able to train with batch sizes up-to 200. While this is lower than the other models, it is still significantly high enough to get good training convergence. However, with BART-large this number reduces significantly to around 50. This might be one of the reasons why we do not get similar improvements with BART-Large models. Since the training complexity of HNet-MAML is higher compared to the other models, we only train the model for 10 epochs. In contrast, we train the other models for 20 epochs.
    
    \begin{figure}
    	\centering
    	\includegraphics[scale=0.3,trim={0 240 0 80},clip]{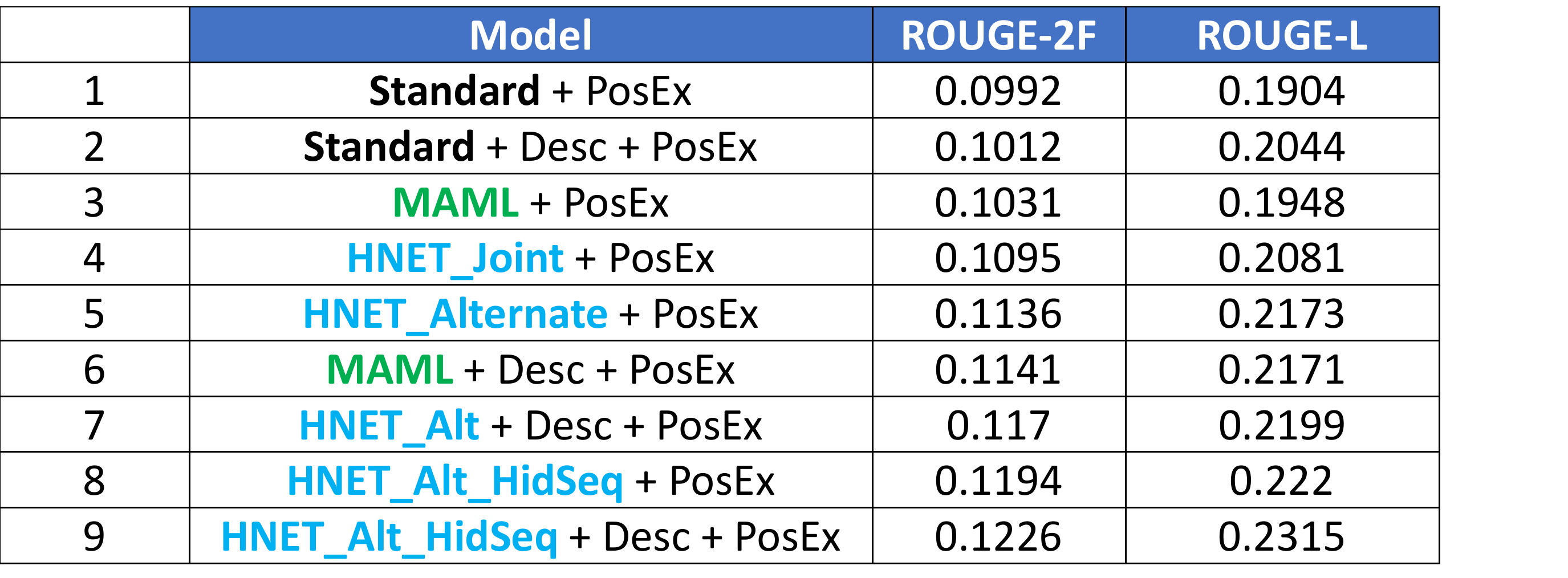}
    	\caption{Ablation studies with on the summarization/title-generation validation set for different models. The proposed model enhancements improve the performance.}
    	\label{fig:ShortAblationsTable}
    \end{figure}

    \begin{figure}
    	\centering
    	\includegraphics[scale=0.27,trim={0 100 0 80},clip]{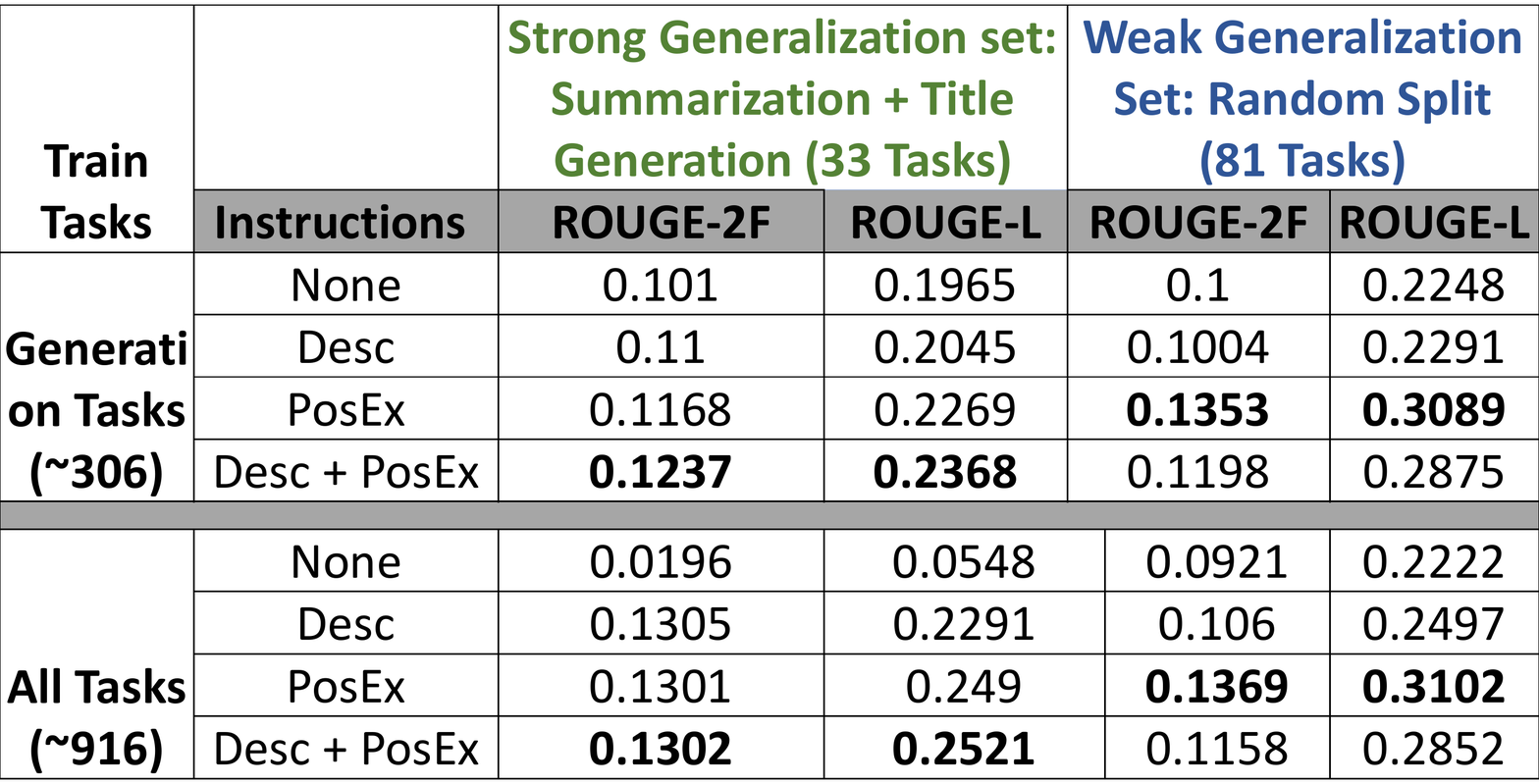}
    	\caption{Baseline metrics with standard training and different instructional fields. Instructions improve the performance over the \textit{None} setting. Best metrics are obtained when using Desc + PosEx.}
    	\label{fig:StandardAblationsTable}
    \end{figure}
    
    \section{Experimental settings}
    In the main section of the paper we have summarized the settings for our experiments. Here we provide some more details. 
    
    We use the BART encoder for both the HNet and the Main-LM models. We use a maximum context length of 1024 for the encoder and 128 for the decoder. Note that in MTIL, the inputs to the encoder includes the task description, one positive example and the training instance with an input and output text. This significantly increases the context length of inputs compared to traditional supervised training. In addition, we consider tasks such as summarization, whose inputs are typically long. Thus some of the instances of training and evaluation cannot be fitted into the chosen context windows. 
    
    We do not truncate the inputs. Instead we remove those instances which exceed the maximum context lengths. I.e. we compute the tokenized lengths of instances with the longest input configuration (Instruction = Desc + Positive Example) and remove all instances which exceed the context lengths of encoder (we do the same for the decoder max length of 128). Thus for all configurations, the training, validation and the evaluation instances remain the same irrespective of the input instruction configurations. Some tasks such as CNN-Daily mail summarization get completely filtered by this process due to the long input as well as the long instructions. For others it is partial, but since we use a maximum of 100 instances per task during training (much lower than total instances available), each task has similar number of instances. For the all tasks training set, we have a total of around 120k train instances, while we have around 52k for the smaller generation task sets.

    \section{Baseline Metrics with Standard Training}

    Here, we  report the baseline performance using the standard training approach, and different instruction components in Figure \ref{fig:StandardAblationsTable}. 
    
    \textbf{Role of task description}: Long task descriptions by themselves are useful ("Desc"  is better than "None") for both train task sets.  Task description has a bigger impact when using the all-tasks set of 916 tasks. This is possibly because the short task description becomes ambiguous (e.g. several question generation tasks may have the same short description) across more number of tasks.
    
    \begin{figure*}
    	\centering
    	\includegraphics[scale=0.65,trim={0 80 0 80},clip]{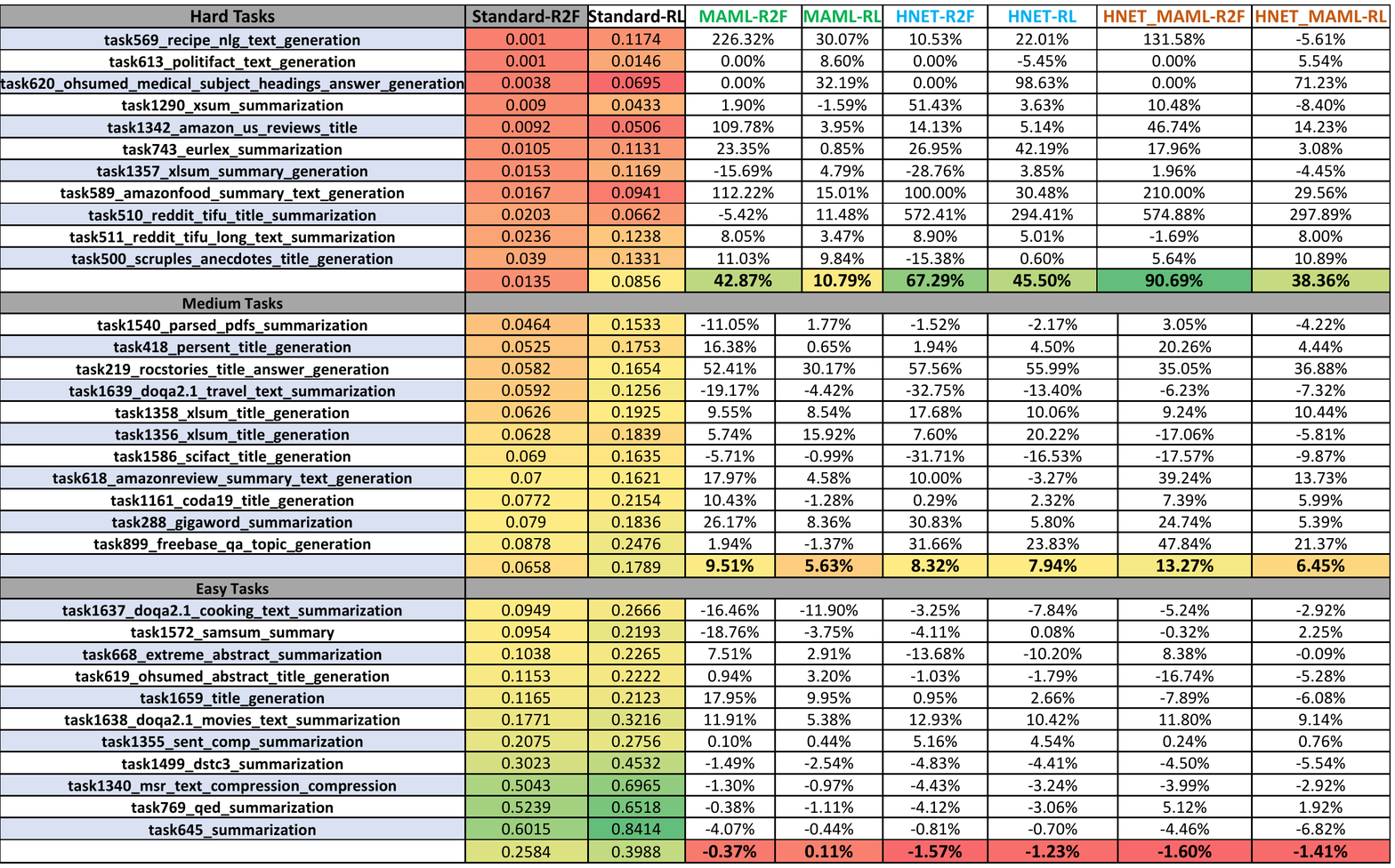}
    	\caption{\% differences listed for individual tasks divided into easy/medium/hard difficulty levels. Results show that MAML, HNet and HNet-MAML models have significant improvements for the difficult tasks)
    	}
    	\label{fig:DetailedTaskLevelDifficultyGenerationTasks_All_Metrics}
    \end{figure*}
    
    \textbf{Weak vs. strong generalization eval set}: Descriptions improve generalization but only for the strong generalization set. 
    For the weak set, the best metrics are achieved by using one positive example. 

    \textbf{Tasks vs number of instances}: Using a larger number of tasks leads to better performance as seen Figure~\ref{fig:StandardAblationsTable} (all tasks set is better than the generation set). The effect is not just due to more data in the all-task set. We tested with similar data sizes between the two training sets and found similar results. 

    \textbf{BART Large}: Performance significantly improves for the the larger LM but the overall trends remain the same.  
    
    \section{Pilot studies with different models}
    We  conducted a pilot study with around 100 train tasks and 10\% of the validation data and compared metrics for different models under different settings and hyper-parameters. Here the aim was to find the ideal hyper-parameters as well as quickly validate if the proposed solutions work as expected, before doing more detailed analysis. Moreover, we wanted to use approximately similar settings for different models as this is likely to be the case in a zero shot setting, where we would not have validation datasets to tune our hyperparameters. 

    

    We summarize the best performance from different models in this pilot study in \ref{fig:ShortAblationsTable}.  We see that using both the task description as well as a positive example has the best performance across all models. For the HNet based model, we see that using the training schedule of alternately freezing the Main and HNet  language model has better performance than jointly training both the networks. We also see that using the sequence of hidden states instead of the last hidden from the HNET decoder leads to better performance as expected. 
    
    \begin{figure*}
    	\centering
    	\includegraphics[scale=1.28,trim={0 80 0 80},clip]{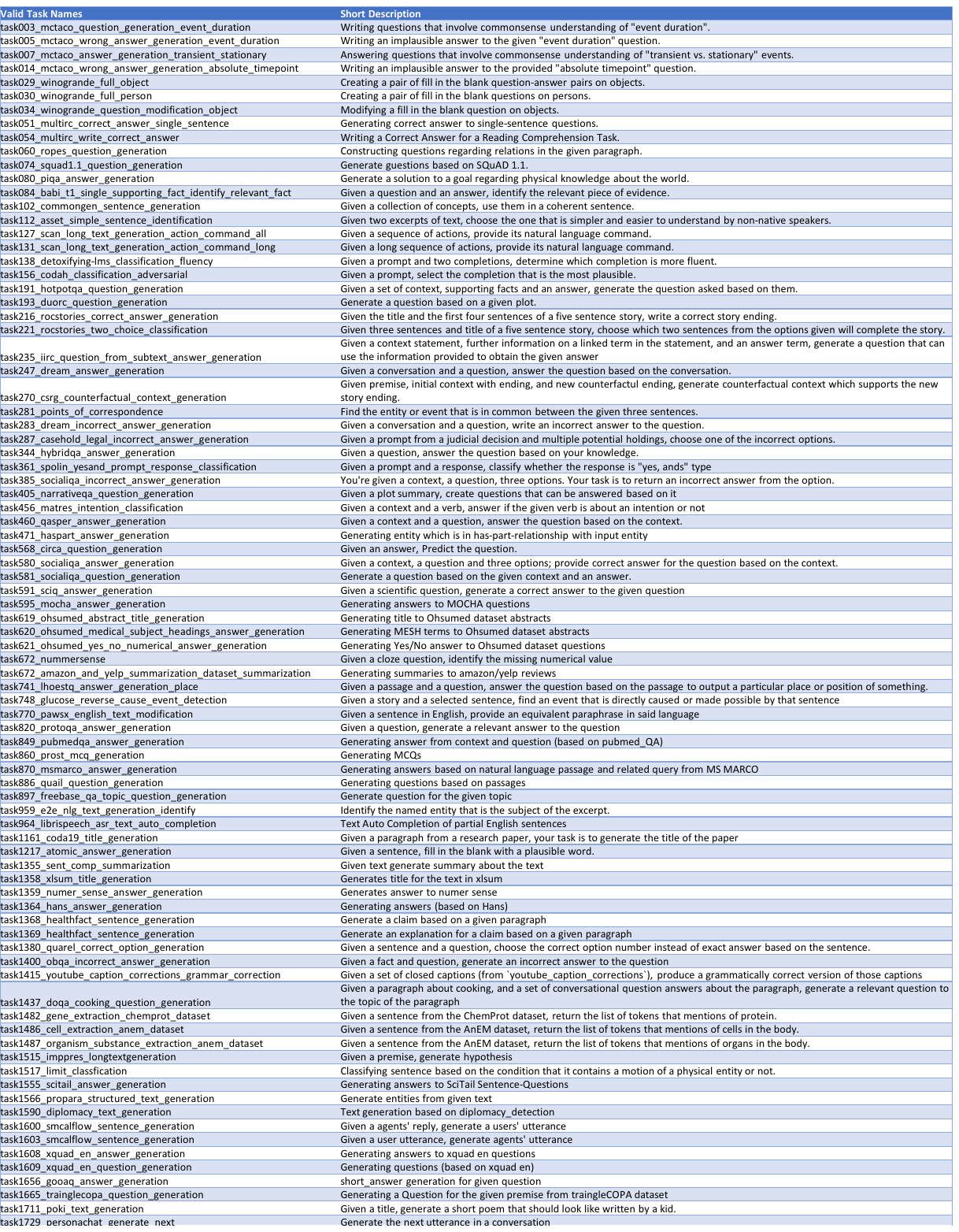}
    	\caption{Weak generalization evaluation set: List of tasks with the short task descriptions for the weak generalization set of 81 generation tasks}
    	\label{fig:WeakGenerationSet}
    \end{figure*}
    
    \begin{figure*}
    	\centering
    	\includegraphics[scale=0.73,trim={0 80 0 80},clip]{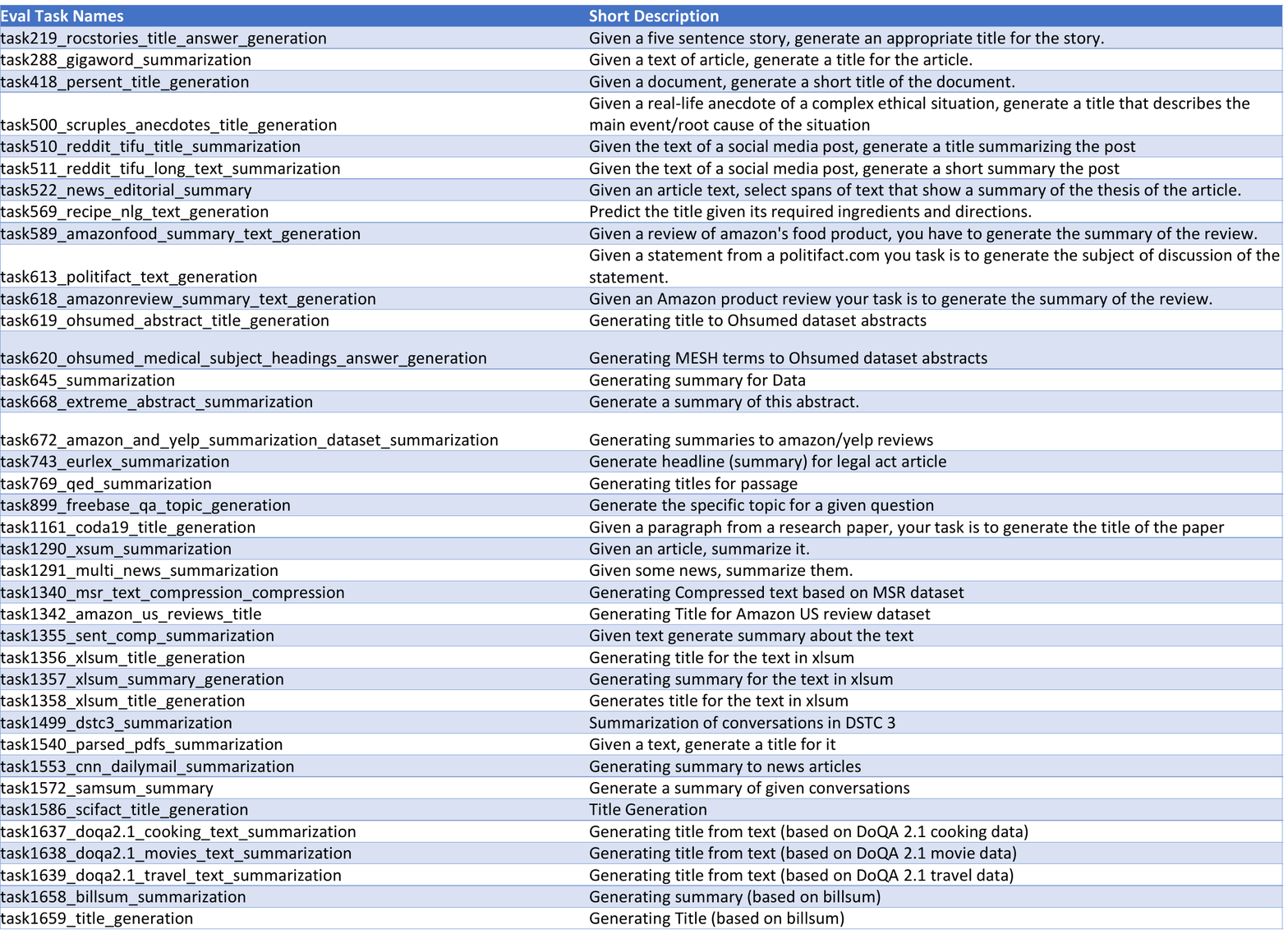}
    	\caption{Strong generalization evaluation set: List of tasks with the short task descriptions for the strong generalization set of 33 generation tasks from summarization and title generation categories}
    	\label{fig:StrongGenerationSet}
    \end{figure*}
    \end{appendix}
\end{document}